\documentclass[
hf]{ceurart}

\usepackage{hyperref}

\newcommand{\name}[1]{\textsc{\textbf{#1}}}
\usepackage{dirtytalk}
\newcommand{\siq}[1]{`#1'}

\usepackage{listings}
\usepackage{csquotes}
\usepackage{url}

\begin{document}

\copyrightyear{2026}
\copyrightclause{Copyright for this paper by its authors.
  Use permitted under Creative Commons License Attribution 4.0
  International (CC BY 4.0).}

\conference{}

\title{Interview with Kalle Lyytinen on ``Implications of Theories of Language for Information Systems''}
\shorttitle{Interview with Kalle Lyytinen}


\author[1]{Kalle J. Lyytinen}[%
orcid=0000-0002-3352-5343,
email=kalle.lyytinen@case.edu,
]
\address[1]{Case Western Reserve University, USA}

\author[2]{Pierre Maier}[%
orcid=0009-0000-4594-6578,
email=pierre.maier@due.de,
]
\cormark[1]
\address[2]{University of Duisburg-Essen, Germany}

\author[3]{Paul Ackah Toffey}[%
orcid=0009-0001-2710-0701,
email=ptoffe1@lsu.edu,
]
\address[3]{Louisiana State University, USA}

\cortext[1]{Corresponding author.}

\begin{abstract}
Over fourty years after the initial publication of \textit{Implications of Theories of Language for Information Systems} in MIS Quaterly, Lyytinen reflects about the origins of his publication and the developments in this area of research over the past decades. In the here presented interview, Lyytinen discusses the linguistic core of information systems also in light of recent trends and developments in the field, especially with regards to large language models and generative AI. Future research directions following a linguistic perspective on Information Systems (IS) research are outlined.

\end{abstract}


\maketitle

\section{Introduction and Editorial Remarks}

This paper presents the contents of an interview conducted with Kalle Lyytinen on his 1985 MISQ article \textit{Implications of Theories of Language for Information Systems} \cite{Lyytinen1985}. A previous version of the article has been presented at the 1984 \textit{International Conference on Information Systems} (ICIS) \cite{Lyytinen1984}. The interview was conducted by Pierre Maier and Paul Ackah Toffey on 7 November 2025 over Zoom. 

In his 1985 article, Lyytinen maintains that language and human communication are inextricably part of information systems. He writes: \say{The very idea of an information system [...] is to provide a means and an environment for human communication. In this sense, information systems [...] would be useless without a linguistic function. Information systems development involves a language development and formalization process. Thus, development processes, methods, and research programs, explicitly or implicitly, are based on a theory of linguistic phenomena} \cite[p. 61]{Lyytinen1985}. The main part of the paper is devoted to presenting, comparing, and discussing implications of five different language views. The five language views presented in the paper are called (1) the Fregean core, (2) the Chomskyan grammar, (3) Piaget's schema, (4) the Skinnerian response, and (5) ordinary speaking. According to Lyytinen, different language views may be adopted in researching information systems, stating that \say{each view entails a different strategy for the analysis and is appropriate for different tasks. [...] A contingency approach is clearly needed to decide which views to adopt, given the development environment} \cite[p. 71]{Lyytinen1985}. 

The interview has been conducted as part of the MISQ Insider initiative. A more comprehensive version of the interview, including also self-presented profiles of all participants, has been published on LinkedIn by MISQ Insider.\footnote{\url{https://www.linkedin.com/feed/update/urn:li:activity:7483223985669758977}, last accessed 24 July 2026} The interview as presented in this paper has been edited for improved legibility and includes references to publications mentioned in the interview.

\section{Main Interview}

\name{Maier:} How did you come to consider the study of different language views, or language in general, to be insightful for IS research?

\name{Lyytinen:} My original interest was in databases and conceptual modeling. That was a hot topic in the early 90s. Entity-relationship modeling emerged in the mid-70s. Modeling got interested more in a logical level, in higher-level concepts for capturing the meaning of data. The same for the famous ISO report on conceptual modeling, which was much of the groundwork at that time. They also talked about different views of the semantics that you capture but they never defined what this \siq{semantic} is. And I was somewhat puzzled about that. I said that there needs to be something deeper to this. I started to read broadly from linguistics, philosophy of language, cognitive linguistics, psychology, and different fields about what they say about meaning and semantics. I found that it's not at all a simple term. It's a very, very complex notion, and we are still struggling with that. It hasn't been sorted out. Contrary to what people in, for example, AI claim. I found out that there are multiple different positions on it, and then I thought that you have these positions, and I also read the literature on IS, and I said that they seem to align with certain of these different positions, and at least somebody has to try to, at least I for myself, I have to try to make sense of this. What are these different views and how do they relate to one another? And that was the sort of origin of the writing of this paper. 

It was originally, I think, meant to be some sort of introduction or a part of the introduction for my thesis. But I changed my thesis topic, so it never became part of the thesis. I actually had a German advisor, the late Heinz Klein. And he said: \say{Conceptual modeling is a bit boring. You have to look at something bigger.} So I looked at social action and went on to that field. But that's pretty much the sort of thesis, why and how I came to look at these different language views, and tried to make sense of them. 

I also wrote some other commentaries. Their bits are on the record. They have been published a little bit later. There's one published in \textit{Information and Management} \cite{Lyytinen1987b} where I debunk fact-based goals and say that there are problems in how you'll constitute facts, that facts are not given, they are socially constructed. You have to understand the processes through which people actually create categories, social categories, which are then a foundation for establishing the meaning of the terms. I also, with Hirscheim and Klein, wrote a book in 1995 \cite{HirschheimEtAl1995} which discusses the philosophical and conceptual foundations of systems development and data modeling.

\name{Toffey:} I would like to ask a question concerning you saying that the \say{history of IS is a history of the growing awareness of linguistic essence} \cite[p. 72]{Lyytinen1985}. How does this hold 40 years later?

\name{Lyytinen:} Well, it's an interesting issue. If you consider it sort of ontologically, or \siq{what information systems are,} they are essentially linguistic systems. I also call them semiotic systems. They're just bit strings. And it is the human who gives these bit strings a certain type of meaning, which is to say that the meaning lies somehow outside the bit strings themselves. If you take a pure Computer Science perspective, the computer is just taking bit strings in, storing bit strings, and putting out bit strings. That's what it does. So the fact that we see that this has some sort of meaningful activity, either in performative terms or in terms of these things, is socially constructed. You cannot negate or avoid the fact that there's a linguistic or semiotic essence in information systems. If you read much of the literature, this is often taken as given. It's like a background, taken for granted. And then you look at other aspects which follow from the fact that that they have linguistic capability or semiotic capability. What do these systems do? How do they have economic effects, or some other types of effects? And then the other string of research looks at it because there's work in manipulating any type of linguistic representation. 

If you go back 100 or 200 years, people were maintaining certain types of records in cabinets and files. It was a lot of human work. And so, much of the computational effects have been actually trying to automate that type of work. But the other aspect is that there is always this linguistic dimension to that. And it cannot be avoided. It can be ignored or it can be taken for granted. There have been fluctuations regarding to what extent people are thinking that this is a central aspect of the IS field or not. If you go to Weber \cite[e.g.,][]{Weber1997}, and other ones, I don't see their view of what these systems \textit{do}. But they have started from the position that the essence of the IS field is largely the fact that they are representational systems, semiotic systems, and we have to build this theory of representations -- which they have developed. That's one part. You can see that they have never been in the mainstream but they have still been recognized. People who work on this topic have always been part of the community. So it has never disappeared. But the awareness of the linguistic essence, I would say that, given the other enormous effects these systems or digital technologies currently have, people often take it for granted. I think there's much more to be gained in trying to understand what it actually means, in terms of the effects of these systems and the ways in which they are being used. 

There are also new innovations and aspects of this, which go beyond the \siq{traditional linguistic essence.} If you look at computers in automated driving systems. It's not really a traditional way of using language, which I was focusing on in the article because that was the only thing where the computers were at that time used for. It is largely trying to simulate and emulate or perform certain aspects of human cognition and action. It's also founded on cognitive and linguistic categories. For example, the driving system needs to recognize that there's a car. \textsc{Car} is a linguistic category, just as \textsc{Bicyclist} or \textsc{Signpost.} These are categories created and defined by linguistic, socially conveyed categories. They are the foundation for building ADAS (advanced driver-assistance systems). The way in which these systems perform goes further beyond. It's not about talking or communicating -- exchanging information -- it's performing certain tasks. Of course, much of our task, which we do, is founded in linguistics. That's what Wittgenstein spoke of as forms of life. You cannot avoid it. 
It comes in higher or lower levels and it's such a complicated phenomenon that there's so many different positions on it.

\name{Toffey:} It seems so. You are saying something that got us into an argument during our first seminar classes concerning whether IS is an IT artifact or not. Regarding Benbasat and Zmud's paper \cite{Benbasat&Zmud2003}, which talks about the identity crisis within the IS discipline and defines its core properties. Do you see IS as an IT discipline or maybe something else? 

\name{Lyytinen:} Well, I also wrote an argument about the identity crisis \cite{Lyytinen&King2004}, and I argued that identity is either defined by theories or by the core issues which the field is obliged to or interested in. Those core issues must be salient, which means that they must matter not only for single scholars but for the larger community -- so that you can train people who can find jobs, or so you can answer issues or topics which are important to outside people like policy makers, business leaders, and so forth. As part of that, you build and develop explanations and look at the phenomena and try to understand them. 

We have always, from the get-go, looked at systems. The scale of these systems and the units of analysis of the types of systems we are interested in, of course, have changed dramatically over time as computerization, what we now call digitization, has expanded. But the systems have three unique essential characteristics. First, they are work systems. There are certain types of work that gets done. Secondly, you have digital technologies, which means that some elements in these work systems are carried out by digital technologies which shuffle bit strings back and forth and store them. These bit strings have to be made meaningful and impactful. If you cannot do that, they don't have any value. You cannot integrate them into the work. Understanding how these work systems are designed, what their effects are, how they need to be organized, how they evolve, and so forth, has been, if you look at the history of IS, the key focus of that. But given the versatility of the way in which digital technologies can be embedded and how the digital capabilities are expanded, the units of analysis, the types of analysis we need to do, the effects, and all this, have changed over time. That's my short answer to that question.

\name{Toffey:} We read your paper \textit{Nothing At The Center?} \cite{Lyytinen&King2004}, I think that's what you are discussing right now. There are people in this IS artifact.

\name{Lyytinen:} There is always an artifact in IS, because work systems are artifacts. The digital artifact is a part of that artifact. Of course, if you say that it is the digital artifact itself, then it's Computer Science. If it's the artifacts and their relationships in the broader social system, and their role in this broader socio-technical system, then it is Information Systems. But the point is that you cannot take the artifact away. It is part of that, and that's part of the system definition. If you look at other fields, like organization theory, strategy, and so forth, they typically black-box the artifact. They treat it as a context or a proxy. If you take the Orlikowski view, it's just the context. We look at IT systems versus some other types of systems. But they don't theorize or characterize or analyze in any way what type of artifact there is, other than labeling it somehow, or looking at its cost, or something. And the role of IS has always been to understand the functions of these systems, or the capabilities, and how they influence the way in which the system is configured and how it performs. Now that idea is, of course, reflected in our ever-expanding and also everlasting exercise on affordances. It's just one way of trying to capture the relations between the use and the user, and the other side has to be performative.

\name{Maier:} In your paper, you suggested that we should focus on multiple language views and try to integrate them instead of focusing on one exclusively, which, according to your observation, was done by many publications at the time. Is it your impression that 40 years after publication this suggestion has been followed by researchers or practitioners?

\name{Lyytinen:} No. I don't think so, because it's difficult. Let me try to clarify. If you're a designer of a system or somebody who has to deal with building, maintaining, or evaluating certain types of systems -- IT artifacts -- these people need to do so based on their own understanding and competencies in language. They need to, in order to evaluate this system, be able to shift between these different language views, at least at some level. So if you are a practicing systems developer or IT manager, you intuitively are dealing with these different views, different settings, different problems, different questions. And I, of course, was very naïve at that time and also very, I would say, sort of Cartesian, and had the idea that you design a method and you follow that method, and then it will automatically result in a better system. I thought that if you can systematically classify these views, integrate them into what you need to do, and which type of questions they lead to ask, and what types of explanations they provide, you could develop a much more encompassing, systematic way of building systems. Nowadays, it looks like systems are built in a totally different way, but you integrate certain aspects of that intuitively or unconsciously into these design methods. Like, if you think of agile development methods, which are based on very fast feedback-based learning. In principle, they say that we cannot ex-ante determine what the impacts are of using this artifact in the way in which it conveys certain meanings in a certain setting. We cannot figure it out by asking people. Basically, it says \say{let's try to figure out what other effects it may have and how people react to that system and how they reason with it,} and they use that feedback as a way of developing that. So they use a pragmatic approach and certain conceptual-modeling views, which are intuitively embedded into that, but in a very different way of developing the systems. And of course, the syntactic part has always been there. You have to define syntax. Although certain aspects of the syntax are now much easier than before, because the technology has advanced. So some of the arguments we made there probably are not that relevant now as they were at that point in time. But it also shows that I made probably incorrect or naïve assumptions about the way in which you can actually develop these types of digitally-based artifacts.

\name{Maier:} Just to clarify: You think that there are certain influences by different language users that are not clearly theoretically distinguished?

\name{Lyytinen:} Yeah, people haven't followed that. And there are also snippets of that in different parts of the publications in the field over the last 40 to 50 years. For each of these views, you can certainly, if you were to do some sort of systematic literature review, find that there's a stream of work which has been carried out on that, and we know much more about it now. The view which I had at that time reflected the idea that we had about information systems in 1985, but it's pretty different from what it is now. At that time, just to give you a sense, although it wasn't that archaic, office automation was becoming a hot topic. So, how do you integrate Excel and email and all that together with networking. These were the hot topics.

\name{Maier:} Then maybe we need new, updated language views.

\name{Lyytinen:} Yes.

\name{Toffey:} In your paper, you highlighted that requirements analysis serves as a key application area of the different language views. How do different language views affect not just requirements analysis, but also, for example, implementation and testing? 

\name{Lyytinen:} Well, I can't remember. This is a part of the paper I cannot fully recall. If you look at, for example, testing. Testing is a syntactic exercise and it has always been. You keep certain test cases and provide syntactic inputs. You look at whether the system is able to produce correct outputs as stated in the system specification. It's a syntactic exercise. The people who are doing the testing are not really interested in what the things mean. They may have a certain idea of what these are supposed to mean but it is basically based on syntactic exercises. It's relatively straightforward and simple in terms of the basic idea of testing. Now, of course, testing as an activity, if you find bugs, is very complicated, because you have to go and change the code. And the technical system is complicated. But the basic idea is based on this syntactic idea. And in a way, you could also say that it is founded on this idea of generative grammar. You say that bit string types are combinations of data, or different types of data, and which combinations in certain sequences are acceptable and which are not. It's very close to this idea of a generative language that you can generate. You have a certain variance in how you can generate inputs, which is like an opportunity in grammar for that, and similarly, you have a certain variance in how you produce the outputs. And you have to figure out that these inputs and outputs are as defined. 

In the implementation, of course, there are certain aspects which are beyond just a sort of linguistic view. This is why you look at the technical -- what people generally call the non-functional -- aspects of the system which are just as important, like, response speed, reliability, how buggy the software is, or the error rates in it.

\name{Maier:} What I was also thinking with regards to that question is that if we are to reconstruct the language of a domain, maybe following the theoretical lenses you provided, we ultimately implement that system using a programming language, like Java, Python, or C\#. And so there are maybe differences in this non-functional regard, what you were saying with performance and typing theories and stuff like that, but these programming languages also provide us with the concepts and the language with which we are to reconstruct the organizational language.

\name{Lyytinen:} Well, that was the original idea I had when I started to work. I use the term professional language, like an accounting language, or an inventory language. You have a certain language to talk about inventories. You have a certain language to talk about accounting systems. You have a certain language to talk about customers. And these languages were already there. It wasn't something that was invented. These were specific professional languages associated with certain functions of the organizations and clear tasks. And these languages were necessary in order to carry out those tasks. So, in Wittgenstein’s terms, they reflect each other in forms of life, which is how you get things done with the language as a community of users or language speakers, so to speak. The original forms of computerization or developing IT artifacts were largely to take those languages, formalize certain parts of them, and embed them in computer systems so that you could use them more effectively. Most of the original aims were automation efforts. You make working with this language much easier. For example, doing inventory is not easy. It's very error-prone. Same with accounting. That was the original reason why people started to do that. And that’s what largely drove the overall use of computing systems, I would say, in the early 90s. The emergence of networks and the emergence of PCs. And then later on, the internet really changed that, because then you would actually move around and carry out all types of different other forms of linguistic communication, like email. It's not a surprise, it's an email. It is an electronic form of sending traditional mail letters. That was the original idea. You just digitized it in a way. Of course, social media is totally different. Then we get into the area where there is no equivalent in the non-computerized space because non-computerized systems couldn't enable these types of activities, or these types of computational processes with linguistic representations, which we have with social media and now with AI systems.

\name{Maier:} In how far do you think can conceptual modeling, with all of its sub-areas like data modeling or business-process modeling, act as an instrument in order to help analyze and reconstruct the language – the professional languages, as you call them – to help account for these different language views? What do you think is the role of conceptual modeling in relation to your work that you presented in that paper?

\name{Lyytinen:} Conceptual modeling, I think, is still highly important and critical, although people don't necessarily understand it. And we don't really talk about it very much. But I think that the way we need to think about conceptual modeling is very different than what I thought in the 1980s, and what a lot of the people who work in conceptual modeling actually think about it. Because much of the work in conceptual modeling at that time, and much of the literature which I followed, looks as if you start modeling a certain domain in a greenfield situation. Of course, there is a designer who models the domain. The conceptual model is basically some type of abstract representation of the grammar, so what you can talk about in that domain and how you can talk about it. But if you look at what actually has happened, it has been up for a long, long time. Like, enterprise ERP systems. They already come with their grammar and language, they are already there. The issue is more or less to what extent you can make the organization adopt that grammar and the concepts, so the semantics of that model and to what extent they are aligned. That's what the matching problems, or the gap problems in ERP development are about. There are some other process flows. Similarly, if you look at business processes, they're a relatively simple way in which you describe the sort of process in which people do it. That's still describing it in linguistic terms but it's not a purely linguistic model. It's basically task modeling. But you have to connect the tasks and what people target in those tasks together. That's the idea. That's one thing.

The other one is that you have much higher-level reference models, which are used in order to standardize and coordinate activities between different organizations and different settings. You need to have higher-level models. So there are multiple other types of roles and functions with concepts or models. The way in which you're engaging in conceptual modeling actually enables the design and use of these systems and the integration of these systems. If you look at it now, micro-level architectures and related ideas about microservices are being discussed. Microservices basically come with a certain conceptual model behind them. It’s evolving all the time, a little bit, but it is a relatively stable conceptual model. How will payment be done and so forth. And if you look at what conceptual modeling now is, you have sort of separated modeling all these different aspects, and then you develop this orchestration, understanding semantic subtleties in the process. So again, there is a certain aspect of using concepts and engaging in conceptual modeling, but a totally different way than we originally thought about it. And of course, there's a fair amount of work in certain areas which are important, like, reference models and so forth. This led to high-level standards at the firm level or at the even industry level. These are also examples of conceptual modeling. But people don't necessarily use them. Think about it as an exercise -- it's all conceptual modeling.

\name{Maier:} Following your argument, would you say that research needs to put more emphasis on conceptual modeling, or maybe reinvestigate the classic literature on conceptual modeling for current trends?

\name{Lyytinen:} I think that in order to articulate a significant space, it would probably need to be more empirically driven. There's some interesting work that is empirically driven. What is the role and real impact of these types of ways of using conceptual models in different settings? So it's not just about going and deriving the initial model and checking that it is consistent with what you have. There are many more alternative ways in which these models are used. There are also political dimensions to that. There's a fight over the meaning of terms. It's a political game. And organizations do that, industry-level actors do that, and so on. The higher issues about conceptual models have gone, the more political these processes have also become. And people studying conceptual modeling should be aware of that and try to understand also those aspects.

\name{Toffey:} You classified in your paper multiple ways to understand language and showed how each understanding leads to different IS design evaluations. How may your suggested language views guide research on large language models?

\name{Lyytinen:} Well, I think that sort of view, which underlies the large-scale language models, was not considered in that original list to be honest. There wasn't really any literature or work looking at that. Of course, there was, at that time, already some work on linguistics, which looked at statistical distributions of terms in sentences and things like computational linguistics, but it was in a very, I would say, nascent field. There weren’t many interesting things. It was nothing like that you can actually, by analyzing very large datasets, which represent text which are linguistic representations, derive new \siq{cognitive substrates,} higher-level ways in aggregating the meanings. Or rather, certain types of meanings. These are not the same types of meanings that humans attribute. The interesting aspect of those meanings is that the validity of those things is largely determined by us. The output from large language models makes sense if we decide it makes sense. So we are the final arbitrator: are they meaningful or not? The systems themselves don't have a clue what the meaning is. They have just a statistical, very fine-tuned, high-layered model of correlations, which are in these enormous parameter tables. They don't have any concept of meaning. The meaning is in the relationship to us. But they represent a totally new way of analyzing linguistic representations and deriving capabilities to interact in linguistic means with digital artifacts in ways we have never been able to do before. In essence, it's a novel phenomenon. It's an additional, a different way of thinking about how to analyze language.

\name{Maier:} How, if you can expand on that, do you think we, we as an IS community or we as PhD students, should deal with this new phenomenon from a linguistic point of view?

\name{Lyytinen:} As I said, it is an interesting new way of analyzing language and linguistic representations. The other interesting aspect is that it also allows this, which they call multi-modal interaction. So you can integrate visual representations and text-based representations, for example. Any type of representation, sets of types of representations can, in principle, be integrated in some ways during the analysis. And then you can create models which can respond to certain types of queries about that, prompts, as you call them, and provide certain types of responses which are sometimes interesting, and many times they are also meaningful. So they can, for example, provide linguistic accounts of what they can see in a figure. That's one example. Vice versa, they can translate a text into a figure. We didn't have those types of capabilities before because of the sort of universal versatility: anything which is in bit strings can be analyzed in terms of their stochastic properties to higher-level substrates, which they call the cognitive substrate.

\name{Maier:} Do you think we could use LLMs to improve our language which we are using? Maybe even improve the professional languages?

\name{Lyytinen:} Yes, of course, we can. There are other aspects. If you look at Noam Chomsky, he wrote a very sharp criticism of large language models \cite{ChomskyEtAl2023}. People will say that he's an old-school guy because he was sort of behind the whole school of trying to make sense of language, which was sort of using formal grammar models to predict and state whether sentences are correct or not. It's just a very different way and it led to the semantic composition of sentences, which was sort of dominant in the 60s, 70s, and 80s when I did this paper. But he said -- and he still outlines -- that these models don't have any sense of the structure of the language. They don't have any sense of the world. No sense of the semantics of the language, which is the main focus of conceptual modeling, by the way. That was how we tried to do that. That there's a sort of higher-level meaning around, where we can say that these representations are equivalent. Although they have different linguistic representations, which is an important aspect of language. Of course, the stochastic models can do that, too. They can match the high correlations as these things appear to be used exactly in the same place in the sentence. So they probably mean the same thing. That's how they operate, but it's a stochastic, not a sort of world-based analysis of that. It is different.

\name{Maier:} Yes, it's different. I think that these language models lack a normative view of the world. They lack what language should be.

\name{Lyytinen:} They don't have this type of normative aspect at all. Of course, if you look at Chomsky's view, it was normative. It started from the notion that if you give a certain sentence in, like, English, or a formal language, you can answer the question: Is this sentence an acceptable sentence in that language? These models don't make any assumptions about that. Nothing whatsoever. And then, of course, because they are totally descriptive, that's the reason why when you look at how good the model is, the first question is \say{how good was the data which you used to feed into that model?,} because the data determines whether it can really predict whether the language or whether the sentence is correct. If you have just the data which has well-parsed German sentences or English sentences, it will more likely produce good sentences if you have a large enough data set. If you input any type of random stuff, it will generate any type of random stuff.

Let me say one more thing. I think the important aspect of a large language model is that if you think of the general IT stack -- like the device layer, the network layer, the data layer, and then the service layer -- what LLMs, especially large-scale language models, have introduced is that the data layer is different than in the past. We have new types of artifacts, so to speak, in the data layer, which we didn't have before. Enterprise systems didn't have tokens at all and now tokens are put together into these models. So, in the data layer, you have two types of new artifacts: tokens and models which are different from the traditional definitions, like data types and data definitions. And then there are instances. If you look at the other side, you have sort of the data corpus, whatever raw data you have, then you have tokens, and you have models. On the other side, you have the traditional hierarchy of how you model data. And now the data layers in firm-level applications will be composed of all these. And now, when you have these model APIs, like the protocols to use different models as part of some sort of services, you have a much more complicated data layer. And that's a big challenge. At the enterprise level, it's a big challenge to develop applications which are long-term services. It is shaping the way in which we need to think generally about systems design, use, and implementation. If you look at where we are now in the generative AI field, most of the money, most of the interest has been in NVIDIA, so just chips or energy or generalized models that you can plug in. Which model do you put into your, sort of, firm level, which types of models should you use? These sorts of questions. But there's very little about how you should manage at the enterprise level the tokens, how you should manage the models, their interactions, and their relationships to data. That's a totally new type of activity. Do you use general models or specialized models? How do you integrate them with traditional data on services? These are largely unknown but a lot of it is just trying to do that -- learning by trying currently. It will take a long, long time before we are probably out of the woods here. There's a lot of work to be done and especially research in the IS field to be done here. So, being a PhD student studying that sort of area is an interesting topic. It's different, again. This is an IS question because you have to start looking at these language models for certain types of organizations and tasks when you integrate them into these digital artifacts.

\name{Maier:} Just to clarify again, when you talk about models, you said that language models have this new data layer, and that we have token, and we have models, but here you are not referring to conceptual models here, right?

\name{Lyytinen:} No, no, no. These models don't have concepts. Conceptual models are on the side of the traditional model. Now, there's an interesting aspect to that, as you may know, that there's a fair amount of work on explainable AI because these AI systems are black boxes. And also a fair amount of work on how to use conceptual models, or conceptual graphs, as part of the use of the large-scale language models, either in generating the prompt, or also in interpreting the sort of responses. So you could say that, in some ways, well-developed concepts and models may be, in the future, some sort of mechanism through which you can integrate the traditional IT system data with these large-scale language models.

\name{Maier:} Do you think that we need to rethink how we conduct research when using language models in the sense that language models have this, as you said, lack of explainability, and maybe even lack of generalizability. So, how can we meet standard scientific criteria when we conduct research with large language models?

\name{Lyytinen:} These are big problems because there's a reproducibility issue. If you do qualitative research, we have always known that there are similar problems, and then the question is, in qualitative research: \say{Is reproducibility really the main criterion through which you evaluate it?} I think a lot of qualitative scholars would say no. The key point is that you are transparent about your inferences. Somebody else may come up with the same dataset but to another set of inferences. They may be as interesting, or even more interesting, or more valid. But there are multiple different ways how you're going to interpret qualitative data, because it's not pre-interpreted, like in a formal setting if you do surveys and things like that. You, as a scholar, pre-interpret the domain. But there are, of course, issues. If you are given a dataset and you're using it by yourself and in different rounds the same prompt gives you different results. Then, you are in some sort of trouble. At least you have to do some sort of effort to run it enough times to see what the variance is. Or somehow be able to demonstrate that whatever you come up with at the end, if you use these analyses results and you use them to interpret certain aspects that are founded in a certain understanding, what is the variability? And I know that there's some work currently on this topic.

\name{Maier:} If you were to write a paper on the topic of your original publication again, or a related topic on language views or language theories for IS development, what would you make different?

\name{Lyytinen:} Well, first of all, the field of how to interpret and make sense of language has vastly advanced in the last 40 to 50 years. Not so much in the IS field. But in cognitive science, neuroscience, philosophy of language, and sociolinguistics. So it would be much harder work and it would require a recap of where those fields currently are. Nobody has done it, by the way. So that's one thing, because the sort of way in which you frame the whole issue about how you analyze language would need to be updated. In a way, we are relying on certain positions which I outlined and identified in that paper and they still hold. But their relative importance has declined and changed radically over the years because the research has shifted a lot. Especially when you currently have large data corpora of all types of linguistic expressions in digital form. The forms of analyzing and making sense of all that have changed enormously. That's the first thing. 

The other one is that the types of systems we think about currently, in terms of digital artifacts, are totally different from those from 40 years ago. So these two things would need to be worked out first. And then after that, and seeing how they connect, which I did in that work that would need to be carried out, and also that would probably require much more careful analysis of the different streams of literature. Of course, it could be written in a short article, like the one I did. But doing it well in a short article probably would be much more difficult than it was in my time. It's probably more like an extensive survey or a theoretical review of this field. Nobody has done it. Whether anybody's willing to do the job and wants to do it -- if you want to do it -- that's open. It's not easy, I know that. It requires fluency in terms of being able to read different streams of research to interrelate and analyze them systematically. But I think it would also help the field if that were to be done. 

And there are certain new aspects which need to be recognized. One of them I mentioned already, that we have new types of, if you put it that way, semantic artifacts at the data layer. The key aspect of these artifacts is that they are no longer boxed systems. They are more like capabilities or infrastructure components. That's why my interest lies more in infrastructures and how they enable innovation. Because you still build systems but very little of it is greenfield. A lot of that is using existing ones and then adding, expanding, and orchestrating. How the linguistic views are applied there is somewhat different, as I pointed out in some of my comments. 

Let me say one aspect of the history, because it didn't come up anywhere. I was a doctoral student when this was written. I sent it to ICIS. You can find it from, I think, the 1984 ICIS proceedings \cite{Lyytinen1984}. At that time they said that they would select a certain number of the best papers for potential publication in MIS Quarterly. The MIS Quarterly’s editorial board would do some type of additional review on that and then they would publish it. And that's how we got it published. I didn't originally plan for it to be submitted to MIS Quarterly. I probably would not have even done it because I didn't know whether it would even be possible or whether it even had a chance to be accepted. So it was this serendipity. I sent it to the conference, they picked it up, and it went through one round of review. I had to do some revisions but much less tedious and much less painful than they currently are. It got published, like, within a year after, maybe 18 months after I submitted to ICIS. That doesn't happen anymore, so I was lucky. That was my first MIS Quarterly publication. I thought that publishing in MIS Quaterly is easy. I have learned later on that it's not that easy.

\name{Maier:} Do you have anything to add regarding your article or recent developments, or do you want to provide a comment on the current state of IS?

\name{Lyytinen:} If there’s anything to learn for PhD students, it is that you have to do the necessary work, which includes reading and synthesizing the literature so that you know the field. Otherwise, you are not an expert. But the other additional dimension of that is that in order to advance the field, you have to ask a new question or take a new angle. And based on that angle, you should review the literature and say, \say{how does that literature address that question?} Then you'll probably get something that goes beyond purely knowing the field. If you look at the types of literature reviews, which are based on an integrated theory review, there are some in JIS (\textit{Journal of Information Systems}). They typically start with a newer question. There needs to be something which goes beyond. And, if there's one thing which I learned from this paper which I applied later, is that I sort of learned that you have to ask a sort of higher-level, a little bit more abstract question, which provides an overarching view. When I did my thesis, I asked the questions about how has the systems development literature addressed the question of whether the system succeeds or fails as a way in which you organize the development process. And that led to an article which was published in ACM Computing Surveys in 1987 \cite{Lyytinen1987c}.

\name{Toffey:} I would like to find out if there is any emerging phenomenon within the IS field that, as a PhD student, I can look into currently.

\name{Lyytinen:} Well, there are different ways of doing that. Everybody, of course, follows new technology. I guess that there are a gazillion students currently doing some work on AI. And that's valuable and useful. It's basically because the first thing which you have to do as a PhD student is to show that you can command the craft. You can do research which is rigorous and meets the standards. It's basically demonstrated that you have the competency to do research. That's the first way in which you're allowed to your PhD thesis. The second one is: can you produce something that has credibility? But the second question is, if you ask that, is it novel? Is it interesting? These are the two other questions. Some people have access to a novel phenomenon that nobody else has access to. And then you can claim, based on that, that you have access to the hottest technology. This would be one example. And then what is interesting is largely the question: does it change the way in which we theorize about this phenomenon? Or does it change the way in which we think about, in terms of practice, how we are going to solve or work with this problem? And these two other ones are much harder to do, and as I said, the thing which probably helped me is to do this and bring novelty at that time. It was much easier than now because I asked different questions on that phenomenon, like systems development. I didn't talk just about modeling or something like implementation. I asked a question: What is the connection to success or failure? And what does the literature say about how this reduces the likelihood of failure or increases the likelihood of success if you do X, Y, or Z? And what are the grounds on which they say that there is this connection? I pointed out here, for example, that these language models and also the tokens are a new form of data. It’s a data layer. So, there's a lot of interesting questions we can ask, novel questions we can ask: What are the consequences? How we theorize about data? How we organize the work around it? Because there's work which needs to be organized around it. How do we use those things in different settings? So, it's not just the service design where a lot of interesting questions can be asked. And of course, the interesting aspect is that they've changed our theories about information systems to the point that they are more performative, agentic, and so forth. They are interesting because these things need to be sorted out because, otherwise, you cannot generate value from these systems. The models themselves have zero value. The value emerges from the use. That has been known for a long, long time. But how is it used now, and how do you expand productive ways of using it? It is a big challenge. And it may require, as we know, reorganizing possibly the whole organization. It may change the way we think about expertise, different ways of using it, and how bodies of expertise in different fields evolve. There are a lot of interesting questions -- novel questions -- to be asked.


\bibliography{map-references}

@proceedings{2026,
 year = {2026},
 title = {Workshops@Modellierung 2026},
 address = {Bonn},
 publisher = {{Gesellschaft f{\"u}r Informatik e.V.}}
}

@article{Benbasat&Zmud2003,
 author = {Benbasat, Izak and Zmud, Robert W.},
 year = {2003},
 title = {The Identity Crisis Within the IS Discipline: Defining and Communicating the Discipline's Core Properties},
 pages = {183--194},
 volume = {27},
 number = {2},
 journal = {MIS Quaterly}
}

@article{ChomskyEtAl2023,
 author = {Chomsky, Noam and Roberts, Ian and Watumull, Jeffrey},
 year = {2023},
 title = {The False Promise of ChatGPT},
 url = {https://www.nytimes.com/2023/03/08/opinion/noam-chomsky-chatgpt-ai.html},
 urldate = {12.11.2025},
 journal = {The New York Times}
}

@book{HirschheimEtAl1995,
 author = {Hirschheim, Rudy and Klein, Heinz K. and Lyytinen, Kalle J.},
 year = {1995},
 title = {Information Systems Development and Data Modeling: Conceptual and Philosophical Foundations},
 address = {Cambridge},
 publisher = {{Cambridge University Press}}
}

@article{Lyytinen&King2004,
 author = {Lyytinen, Kalle J. and King, John Leslie},
 year = {2004},
 title = {Nothing At the Center?: Academic Legitimacy in the Information Systems Field},
 pages = {220--246},
 volume = {5},
 number = {6},
 journal = {Journal of the Association for Information Systems}
}

@inproceedings{Lyytinen1984,
 author = {Lyytinen, Kalle J.},
 title = {Theories of Language and Information Systems: An Appraisal of Alternative Language Views for Information Systems},
 pages = {149--150},
 booktitle = {ICIS 1984 Proceedings},
 year = {1984}
}

@article{Lyytinen1985,
 author = {Lyytinen, Kalle J.},
 year = {1985},
 title = {Implications of Theories of Language for Information Systems},
 pages = {61--74},
 volume = {9},
 number = {1},
 journal = {MIS Quaterly}
}

@article{Lyytinen1987b,
 author = {Lyytinen, Kalle J.},
 year = {1987},
 title = {Two Views on Information Modeling},
 pages = {9--19},
 volume = {12},
 number = {1},
 journal = {Information {\&} Management}
}

@article{Lyytinen1987c,
 author = {Lyytinen, Kalle J.},
 year = {1987},
 title = {Different Perspectives on Information Systems: Problems and Solutions},
 pages = {5--46},
 volume = {19},
 number = {1},
 journal = {ACM Computing Surveys}
}

@book{Weber1997,
 author = {Weber, Ron},
 year = {1997},
 title = {Ontological Foundations of Information Systems},
 address = {Melbourne},
 publisher = {{Coppers {\&} Lybrand}}
}

\end{document}